\documentclass[journal]{IEEEtran}

\usepackage{epsfig}
\usepackage{graphicx}
\usepackage{subfigure}
\usepackage{amsmath,bm}
\usepackage{amsfonts}
\usepackage{cite}
\usepackage{diagbox}
\usepackage{multirow}
\usepackage{amsthm}
\usepackage[utf8]{inputenc}
\usepackage{algorithmic}
\usepackage{algorithm}
\usepackage{threeparttable}
\usepackage{color,soul}

\newtheorem{definition}{Definition}
\ifCLASSINFOpdf
\else
\fi
\begin{document}
\title{Predicting Real-Time Locational Marginal Prices: A GAN-Based Video Prediction Approach}

\author{Zhongxia~Zhang,~\IEEEmembership{Student~Member,~IEEE,}
	and~Meng~Wu,~\IEEEmembership{Member,~IEEE}
	\thanks{Zhongxia Zhang and Meng Wu are with the School of Electrical, Computer and Energy Engineering, Arizona State University, Tempe, AZ, 85281 USA email: zzhan300@asu.edu, mwu@asu.edu.}% <-this % stops a space
	
	\thanks{}}

\maketitle

\begin{abstract}
In this paper, we propose an unsupervised data-driven approach to predict real-time locational marginal prices (RTLMPs). The proposed approach is built upon a general data structure for organizing system-wide heterogeneous market data streams into the format of market data images and videos. Leveraging this general data structure, the system-wide RTLMP prediction problem is formulated as a video prediction problem. A video prediction model based on generative adversarial networks (GAN) is proposed to learn the spatio-temporal correlations among historical RTLMPs and predict system-wide RTLMPs for the next hour. An autoregressive moving average (ARMA) calibration method is adopted to improve the prediction accuracy. The proposed RTLMP prediction method takes public market data as inputs, without requiring any confidential information on system topology, model parameters, or market operating details. Case studies using public market data from ISO New England (ISO-NE) and Southwest Power Pool (SPP) demonstrate that the proposed method is able to learn spatio-temporal correlations among RTLMPs and perform accurate RTLMP prediction.
\end{abstract}


% \fi
\IEEEpeerreviewmaketitle

\section{Introduction}
\IEEEPARstart{L}{ocational} marginal price (LMP) prediction is critical for energy market participants to develop optimal bidding strategies and maximize their profits. However, the increasing integration of renewable resources leads to high uncertainties in both electricity supply and demand, which then increases price volatility in electricity markets and makes LMP prediction difficult for market participants.

To mitigate price volatility, LMPs in US electricity markets are settled twice in the day-ahead (DA) and real-time (RT) markets\cite{rule1,pjm}. As a spot market, the RT market experiences higher price volatility compared to the DA market (which is a forward market) \cite{nla.cat-vn153514}. Therefore, real-time LMPs (RTLMPs) are less predictable compared to day-ahead LMPs (DALMPs).

LMPs can be accurately predicted if the power grid parameters, topology, and operating conditions are perfectly known. However, these confidential modeling details are typically not shared with market participants. This leads to more challenges for market participants to predict the highly volatile RTLMPs.

%it is challenging for market participants to accurately predict LMPs, especially RTLMPs with high price volatility. This paper focuses on the prediction of system-wide RTLMPs.

% are determined based on physical network parameters, topology, and system operating conditions. Since these modeling details are kept confidential by system operators, it is difficult for market participants to predict LMPs without these confidential system details. 

Existing methods predict LMPs from either system operators' perspective or market participants' perspective. In \cite{1294980,773811,deng2016probabilistic}, simulation-based methods and multiparametric programming approaches are proposed to predict LMPs from system operators' perspective. These methods assume perfect knowledge of system models, which is not available to market participants. In \cite{7285827,7514918,7478156}, data-driven LMP prediction algorithms are developed based on the concept of system pattern regions (SPRs) \cite{7285827}. These algorithms do not require detailed information on system operating conditions and network models. In \cite{7478156}, a support vector machine is trained using historical market data to learn the SRPs which represent the relationship between LMPs and system loading conditions. However, this method only predicts future LMP ranges instead of specific LMP values. Moreover, this SPR-based method require predicted nodal load data as inputs for accurate LMP prediction, which is not always available to the public. The method is tested in simulated markets without generation offer variation. These simulations cannot fully represent real-world market operations, since generation offers vary significantly and are highly related to LMP variations in practical energy markets.

From the market participants' perspective, most LMP prediction methods focus on recovering grid models using historical market data \cite{7226869}. These methods estimate parameters of the optimal power flow (OPF) models \cite{Inverse} and capture the underlying market structure of the dc OPF problem \cite{8733097}. These OPF-related methods are computationally expensive. Moreover, their LMP prediction accuracies are highly dependent on the accuracies of the estimated OPF model parameters, which could not be guaranteed for real-world systems with a large number of model parameters that could change over time.

Another group of methods predict LMPs by learning the linear or nonlinear relationship between historical LMPs and uncertain demands. In \cite{7917305,8274124,1425583}, time-series statistical methods, such as ARMAX model\cite{7917305}, ARIMA model \cite{8274124}, and AGARCH model \cite{1425583}, are developed to learn such linear relationship. In \cite{1372802,GRNN,4282040}, data-driven machine learning approaches, such as neural network methods \cite{1372802,GRNN} and nearest neighbors model \cite{4282040}, are applied to learn such nonlinear relationship. These methods only capture the temporal correlations between LMPs and demands, without considering the spatial correlations among system-wide LMPs.

To overcome the above disadvantages, we propose a purely data-driven approach for predicting system-wide RTLMPs using spatio-temporal correlations among heterogeneous market data, without requiring any confidential information on system topology, model parameters, or market operating details. In this paper, we make the following contributions:

\begin{enumerate}
	\item We introduce a general data structure for organizing system-wide heterogeneous market data streams (i.e., LMPs, loads, generation offers, weather conditions, etc.) into the format of market data images and videos. This general data structure enables us to leverage various image and video prediction/processing methods for LMP prediction and other data-driven energy market analytics.
	\item We formulate the RTLMP prediction problem as a video prediction problem. A GAN-based \cite{goodfellow2014generative} video prediction model is trained with multiple loss functions to learn the spatio-temporal correlations among historical RTLMPs. This model is applied to predict system-wide RTLMPs for the next hour, using only public data. An ARMA calibration method is proposed to improve the RTLMP prediction accuracy. To our best knowledge, this is the first study incorporating video prediction technology to the prediction and analytics of energy market data.
\end{enumerate}

The rest of the paper is organized as follows: Section II defines the data structures for market data images and videos, and formulates the price prediction problem as a video prediction problem; Section III proposes the GAN-based price prediction model with multiple loss functions; Section IV proposes the ARMA calibration model for improving price prediction accuracy; Section V presents the features selected for the price prediction problem; Section VI presents the case study results; Section VII concludes this paper. 

\section{Problem Formulation}
In this section, we introduce the basic concepts of digital image and video. We then organize system-wide historical market data into a series of time-stamped matrices and map these matrices to a series of market data images. By concatenating these market data images, we form a historical market data video. We formulate the next-hour RTLMP prediction problem as a problem of predicting the next frame of the market data video, given the input historical market data video. With this formulation, state-of-the-art video prediction models can be applied to predict next-hour RTLMPs across the entire system.

\subsection{The Motivating Example: Visualizing LMPs as An Image}
\begin{figure}[!h]
	\centering
	\includegraphics[height=5.3cm,width=6.5cm]{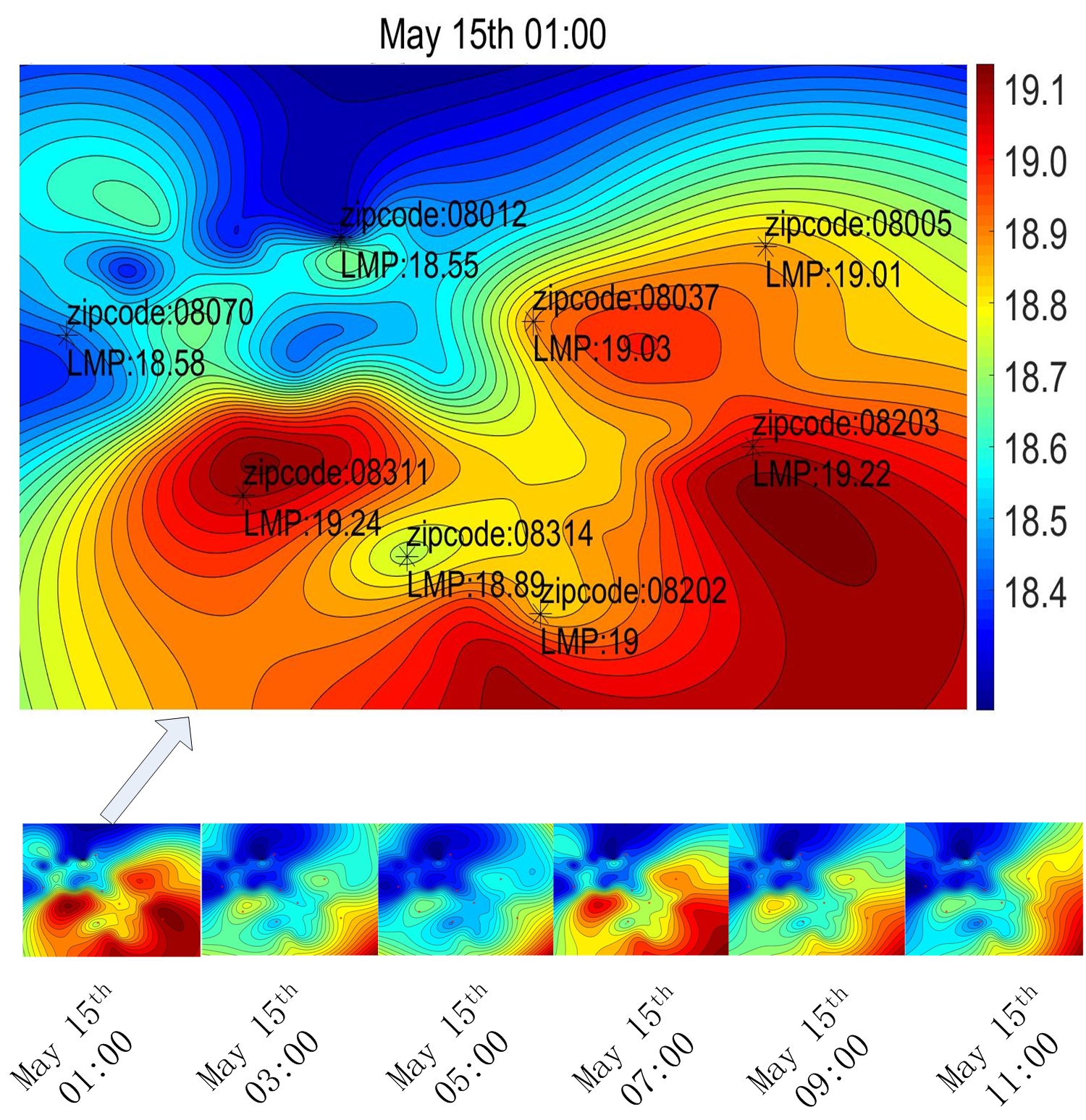} 
	\caption{RTLMP visualization of AECO price zone in PJM market from 1:00 AM to 11:00 AM on May 15, 2019.}
	\label{contour}
\end{figure}

Our LMP prediction method is motivated by the visualization of system-wide LMPs using color-coded values on the corresponding system map\cite{PJM_map}. Fig.~\ref{contour} shows such  visualizations generated using RTLMPs (from 1:00 AM to 11:00 AM on 5/15/2019) obtained from 56 price nodes within the territory of Atlantic Electric Power Company (i.e., the AECO price zone) in PJM Interconnection\cite{PJM_map}. In Fig.~\ref{contour}, eight price nodes are explicitly identified on the image at 1:00 AM with their RTLMP values and corresponding zipcodes. The coordinates of the price nodes represent their geographical locations in the system. At the bottom of Fig.~\ref{contour}, six images are generated using system-wide RTLMPs at six different hours. The colors on these six images vary both spatially (within the same image) and temporally (between consecutive images), indicating the spatio-temporal variations of RTLMPs. This example clearly shows the spatio-temporal correlations among system-wide RTLMPs can be captured by the spatio-temporal color variations across a series of time-stamped images. This series of time-stamped images then forms a video consisting of the above RTLMP visualizations.

%\vspace{-1cm}

\begin{figure}[!h]
	\centering
	\includegraphics[height=2.1cm,width=9.0cm]{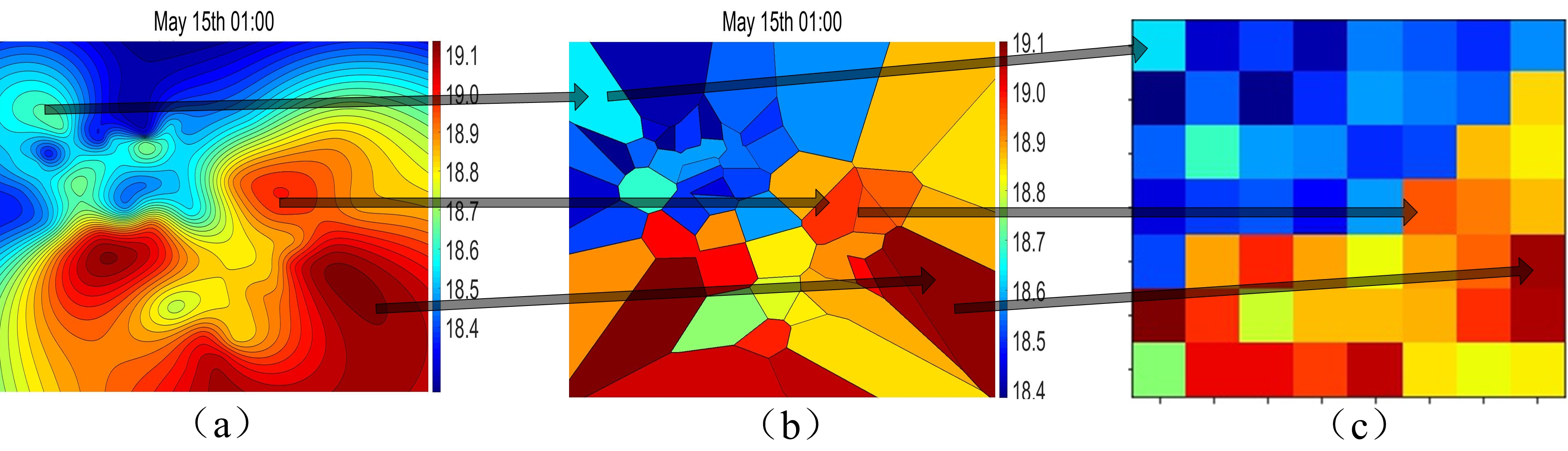} 
	\caption{Comparison between RTLMP visualizations generated using different interpolation techniques.}
	\label{visual compare}
\end{figure}

The smooth RTLMP visualizations in Fig.~\ref{contour} are generated using biharmonic spline interpolation\cite{doi:10.1029/GL014i002p00139}. In Fig.~\ref{visual compare}, two different interpolation techniques (biharmonic spline interpolation for Fig.~\ref{visual compare}(a) and nearest neighbor interpolation for Fig.~\ref{visual compare}(b)) are applied to an identical dataset (RTLMPs at 56 price nodes at 1:00 AM on 5/15/2019, in AECO price node). The nearest neighbor interpolation results in a less smooth RTLMP visualization with exactly 56 different color zones. Each color zone corresponds to the RTLMP value of a particular price node. These 56 color zones in Fig.~\ref{visual compare}(b) are then re-organized to generate the image in Fig.~\ref{visual compare}(c), with 56 colored squares of the same size. The color of each square is fully determined by the RTLMP value at the corresponding price zone. These colored squares represent pixels in a colored digital image, which are the smallest addressable element in an image. In the following section, we formally define data structures for colored digital pixels, images, and videos.

\subsection{Preliminaries: Data Structures for Images and Videos}

%RTLMP maps provided by most ISOs as shown in Fig.~\ref{contour} are generated by fitting a surface which always passes through the data points at given price nodes. Besides the given nodes, the rest of area in the map are interpolated by certain functions. Fig.~\ref{visual compare} shows three different visualizations based on the same color map encoding for the same hour's RTLMPs. Fig.~\ref{visual compare}(a) is general RTLMP contour map fitted by 'V4' function in MATLAB, Fig.~\ref{visual compare}(b) is fitted by 'nearest' function in MATLAB, Fig.~\ref{visual compare}(c) is the digital image we adopt in this paper. In Fig.~\ref{visual compare}(c), each square represents a digital pixel whose color is encoded using a specific price node's RTLMP value. By this format of representation, we can get ride of fitting functions and still keep the spatio-temporal correlations. We introduce the following definitions for colored digital images, pixels, and videos.

%\vspace{-0.5cm}
\begin{definition}
	Let the time-stamped matrix $V(t)$ whose dimension is $M{\times}N$ represent a colored digital image with a resolution of $M{\times}N$. The time stamp $t$ represents the time when the digital image is generated.
\end{definition}

\begin{definition}
	Let the $(i,j)^{th}$ element of matrix $V(t)$ be a $1{\times}3$ vector $v_{i,j}(t)=[v_{i,j}^r(t), v_{i,j}^g(t), v_{i,j}^b(t)]$. This vector defines the $(i,j)^{th}$ pixel of the colored image, where $v_{i,j}^r(t)$, $v_{i,j}^g(t)$, $v_{i,j}^b(t)\in[0,1]$ denote the red, green, and blue color codes for the pixel, respectively. The values of $[v_{i,j}^r(t), v_{i,j}^g(t), v_{i,j}^b(t)]$ indicate the percentages of red, green, and blue color components in the pixel, and fully determine the color intensity of the pixel. For example, $v_{i,j}(t)=[1,0,0]$, $v_{i,j}(t)=[0,1,0]$, $v_{i,j}(t)=[0,0,1]$, $v_{i,j}(t)=[0,0,0]$, and $v_{i,j}(t)=[1,1,1]$ define red, green, blue, black, and white pixels, respectively. 
\end{definition}

\begin{definition}
	Let a series of time-stamped matrices, $V=\{V(1), \cdots, V(t), \cdots, V(T)\}$, represent a digital video obtained during the time interval $[1,T]$. Each time-stamped image $V(t)$ is defined as a frame of the video $V$ at time $t$.
\end{definition}

The above definitions represent a digital video as a series of time-stamped matrices. Each matrix element consists of three real numbers within a certain interval (i.e., the red, green, and blue color codes of a pixel within $[0,1]$). To apply video prediction techniques for our RTLMP prediction problem, in the following section, we normalize historical market data into a certain interval and organize the normalized market data into the data structures of digital images and videos.
%\vspace{-0.3cm}

\subsection{Normalization of Historical Market Data}
Let $C$ be the set of historical market data (of a particular type) collected hourly from $K$ different price nodes within time interval $[1,T]$. Let $c_k(t){\in}C$ be a particular market data point collected from price node $k$ at time $t$, where $k=1,2,\cdots,K$, and $t=1,2,\cdots,T$. For the RTLMP prediction problem, $c_k(t)$ may represent a data point for historical RTLMPs, DALMPs, demands, generations, temperatures, etc. $c_k(t)$ is then normalized to $[-1,1]$ as follows:

%\begin{equation}
%d_k^{norm}(t)=\frac{ln(d_{k}^{+}(t))-ln(max(D^+))/2}{ln(max(D^+))} + \frac{1}{2}
%\label{eqn_norm_1} 
%\end{equation}
%where
%\begin{equation}
%d_{k}^{+}(t)=d_{k}(t)-min(D)+1
%\label{eqn_norm_2}
%\end{equation}
%\begin{equation}
%D^+=\{ d_k^+(t)| k=1,2,\cdots,K \mbox{ and } t=1,2,\cdots,T \}
%\label{eqn_norm_3}
%\end{equation}
%\vspace{-0.3cm}
\begin{equation}
c_k^{norm}(t)=\frac{ln(c_{k}^{+}(t))-ln(max(C^+))/2}{ln(max(C^+)/2}
\label{eqn_norm_1} 
\end{equation}
where
\begin{equation}
c_{k}^{+}(t)=c_{k}(t)-min(C)+1
\label{eqn_norm_2}
\end{equation}
\begin{equation}
C^+=\{c_k^+(t) | k=1,2,\cdots,K; t=1,2,\cdots,T \}
\label{eqn_norm_3}
\end{equation}

In (\ref{eqn_norm_1})-(\ref{eqn_norm_3}), $c_k^{norm}(t)$ denotes the normalized value of $c_{k}(t)$; $min(C)$ denotes the smallest value in $C$; $max(C^+)$ denotes the largest value in $C^+$. The normalized market data $c_k^{norm}(t){\in}[-1,1]$. Although (\ref{eqn_norm_1})-(\ref{eqn_norm_3}) normalize market data to $[-1,1]$, instead of $[0,1]$ (the range of pixel color codes), these two intervals can be easily converted to each other.

%\vspace{-0.3cm}
\subsection{The Market Data Images and Videos}
Consider a wholesale electricity market with $M{\times}N$ price nodes. A set of historical market data of various types (such as RTLMPs, DALMPs, demands, generations, temperatures, etc.) can be collected at each price node. Let $v_{i,j}^{norm}(t) = [v_{i,j}^{norm-1}(t), v_{i,j}^{norm-2}(t), v_{i,j}^{norm-3}(t)]$ be a $1{\times}3$ vector containing three types of normalized market data obtained from the $(i,j)^{th}$ price node at time $t$. Let $V^{norm}(t)$ be a $M{\times}N$ matrix whose $(i,j)^{th}$ element is $v_{i,j}^{norm}(t)$. According to the data structures defined for images and videos, $V^{norm}(t)$ can be viewed as a colored digital image with a resolution of $M{\times}N$, and $v_{i,j}^{norm}(t)$ can be viewed as the $(i,j)^{th}$ pixel of this colored digital image. By concatenating a series of such market data images, we obtain $V^{norm}=\{V^{norm}(1), \cdots, V^{norm}(t), \cdots, V^{norm}(T)\}$, which can be viewed as a colored digital video containing three different types of historical market data obtained at $M{\times}N$ different price nodes for time interval $[1,T]$.

With the above data representation, the correlations among different pixels in the market data images $V^{norm}(t)$, such as $v^{norm}_{i,j}(t), v^{norm}_{i-1,j}(t), v^{norm}_{i,j-1}(t)$, represent the spatial correlations among historical market data collected from different price nodes (locations) at time $t$; the correlations among the $(i,j)^{th}$ pixel obtained at different time, such as $v^{norm}_{i,j}(t-1),v^{norm}_{i,j}(t),v^{norm}_{i,j}(t+1)$, represent the temporal correlations among historical market data collected from the same price node (location) at different time.

\begin{figure}[!h]
	\centering
	\includegraphics[height=4.5cm,width=9.0cm]{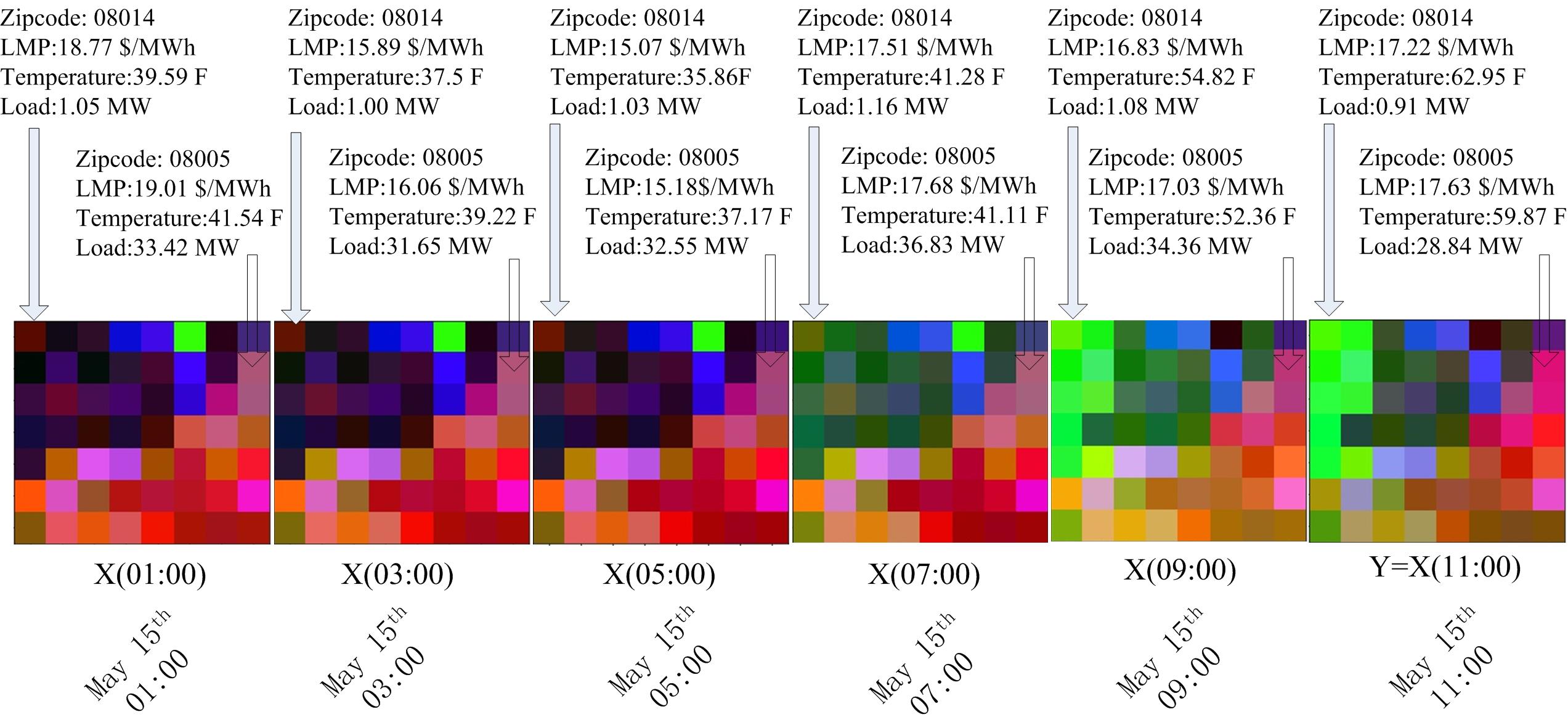} 
	\caption{Colored digital market data video of AECO price zone in PJM market.}
	\label{pixelimage}
\end{figure}

Fig.~\ref{pixelimage} shows a market data video consisting of 6 hourly market data images/frames generated using historical data from 56 (7$\times$8) price nodes in the AECO price zone. Each square in Fig.~\ref{pixelimage} represents a pixel whose red, green, and blue color codes take the value of normalized RTLMP, real power demand, and temperature at the corresponding price node, respectively. The color of each pixel is fully determined by the corresponding market data values (after normalization). It is clear the spatio-temporal variations of the pixel colors represent the spatio-temporal variations of RTLMPs, real power demands, and temperatures during these 6 hours across the AECO price zone. If a learning model is trained to learn the spatio-temporal color variations in the historical market data video, this model could then be applied for system-wide RTLMP predictions.

Although in the above discussion we represent the market data pixel $v_{i,j}^{norm}(t)$ using three types of market data (such as RTLMP, real power demand, and temperature in Fig.~\ref{pixelimage}), this concept of market data pixel can be easily extended to representing $l$ different types of market data, where $l$ is a positive integer. In Section~\ref{feature_selection}, a feature selection process is introduced to identify the $l$ types of market data which contribute the most to the RTLMP prediction problem.

%\vspace{-0.8cm}
\subsection{Formulation of The RTLMP Prediction Problem}
As discussed in the previous section, the normalized historical market data obtained across the system can be organized into a historical market data video $V^{norm}=\{V^{norm}(1), \cdots, V^{norm}(t), \cdots, V^{norm}(T)\}$. The RTLMP prediction problem is then formulated as a video prediction problem. The objective is to generate a future video frame $\hat{V}^{norm}(T+1)$, such that the conditional probability $p(\hat{V}^{norm}(T+1)|V^{norm})$ is maximized. The generated video frame $\hat{V}^{norm}(T+1)$ contains predicted RTLMPs for the future time $T+1$. The predicted market data should follow the spatio-temporal correlations of the historical market data.
%The convincing predicted future image $X^{(T+1)-norm}$ should agree with the spatio-temporal correlations in the historical RTLMP (market data) video.\\
%\indent Given the training data set of normalized historical RTLMPs (market) video $X^{norm}=\{X^{1-norm},\cdots,X^{t-norm}\,\cdots,X^{T-norm}\}$, the objective of this paper is to learn the sptaio-temporal correlations among the normalized pixels in an unsupervised way. A conditional GAN model including an optimal generator model is accepted in this paper to train and predict. A discriminator is utilized in training process to obtain an optimal set of neural network parameters for the generator model. Given a RTLMPs (market data) video, the optimal generator can generate the next frame $X^{(T+1)-norm}$ following $X^{norm}$. After denormalizing $X^{(T+1)-norm}$, RTLMPs at different price nodes (locations) for future time instant $T+1$ is obtained.
%\vspace{-0.3cm}
\section{GAN-Based RTLMP Prediction}
In this section, a deep convolutional GAN model is proposed to solve the above RTLMP prediction problem. The GAN model is trained using multiple loss functions that are capable of capturing correlations among system-wide historical RTLMPs both spatially and temporally. This work is inspired by the video prediction approaches in \cite{radford2015unsupervised,mathieu2015deep,denton2015deep}.
%This section proposes a deep convolutional GAN model for predicting RTLMPs. %In \hl{[X], we introduce a similar GAN-based price prediction approach. This paper improves the price prediction accuracy by including appropriate loss functions, incorporating various types of training inputs into the proposed data structures, and enhancing the calibration process for RTLMPs generated by the GAN model}. 
%The proposed GAN-based price prediction approach accepts multi-loss function to improve the prediction accuracy. This work is inspired by the video prediction approaches in \cite{radford2015unsupervised,mathieu2015deep,denton2015deep}.
%%\vspace{-0.3cm}
\subsection{The GAN-Based RTLMP Prediction Model}
\begin{figure}[!h]
	\centering
	\includegraphics[height=4.0cm,width=9.0cm]{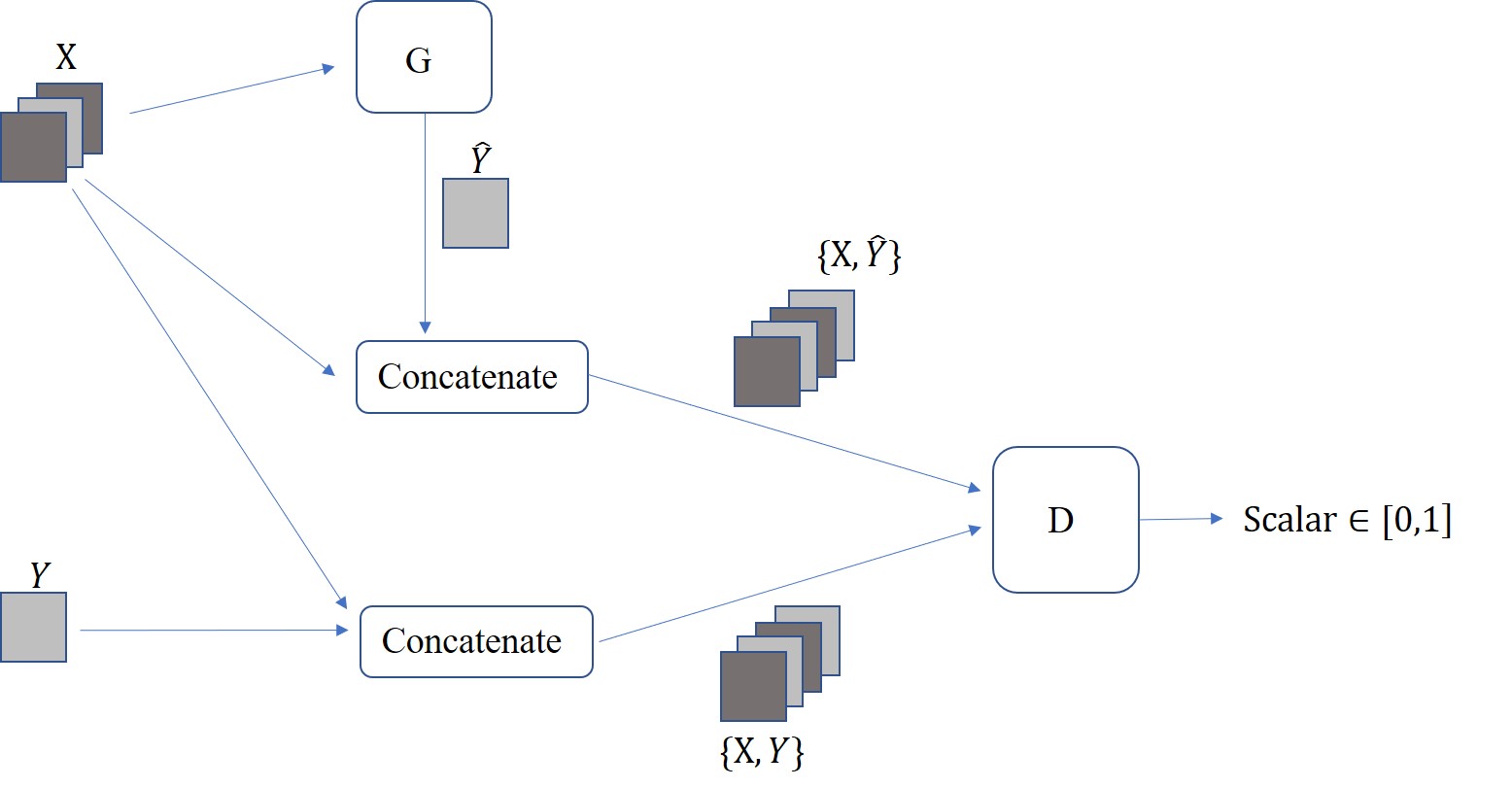} 
	\caption{The training procedure of the GAN-based RTLMP prediction model.}
	\label{training procedure}
\end{figure}

The proposed GAN model consists of a generative model $G$ and a discriminative model $D$. Both $G$ and $D$ are convolutional neural networks. Fig.~\ref{training procedure} shows the training procedure of the GAN model.
%\vspace{-0.3cm}
%\vspace{-0.3cm}
In the architecture shown in Fig.~\ref{training procedure}, $G$ and $D$ denote the generator and discriminator neural networks, respectively; $X=\{X(1),\cdots,X(n)\}$ denotes the input market data video with normalized historical RTLMPs and other types of market data obtained at different price nodes at $n$ consecutive time instants, $X{\subset}V^{norm}$; Y denotes the market data image with normalized historical RTLMPs and other types of market data obtained at different price nodes at time $n+1$, i.e., $Y=X(n+1){\in}V^{norm}$; $\hat{Y}=G(X)$ denotes the generated video frame at time $n+1$, with the predicted RTLMPs at different price nodes. 
%Taking the market data video in Fig.~\ref{pixelimage} as an example, if the above GAN model is applied to predict system-wide RTLMPs at 11:00 AM, given the historical market data at previous hours, while training the GAN model, we have $X=\{X(01:00), X(03:00), X(05:00), X(07:00), X(09:00)\}$ and $Y=X(11:00)$.

%As an example, suppose the above GAN model is applied to predict system-wide RTLMPs
%
% consider the market data video in Fig.~\ref{pixelimage}, input data is a digital video consisting of the first five hourly market data images $X=\{X(01:00), X(03:00), X(05:00), X(07:00), X(09:00)\}$, the future hourly image to predict is $Y=X(11:00)$.
%$Z(\hat{Z})$ denotes the ground truth video (the video with a generated next time image). A scalar between 0 and 1 output by discriminator $D$, indicates the probability of the input video of discriminator $D$ being the ground truth.\\

The generator $G$ is trained to generate a market data image $\hat{Y}=G(X)$ based on the input historical video $X$, such that the conditional probability $P(\hat{Y}|X)$ is maximized. The generated image $\hat{Y}$ and the ground-truth image $Y$ are then concatenated with the input video $X$ to form two new videos, $\{X,\hat{Y}\}$ and $\{X,Y\}$. Taking the new videos as inputs, the discriminator $D$ is then trained to classify $\{X,Y\}$ as real and $\{X,\hat{Y}\}$ as fake. The discriminator output, $D(\{X,\cdot\}){\in}[0,1]$, is a scalar indicating the probability of the input video $\{X,\cdot\}$ being the ground-truth video. Upon training convergence, which means the distance between the generated image $\hat{Y}$ and the ground-truth image $Y$ is small enough given input historical video $X$, the discriminator $D$ cannot classify $\{X,Y\}$ and $\{X,\hat{Y}\}$.
%, $D(\{X,Y\})$ and $D(\{X,\hat{Y}\})$ both approach to 0.5.

%Upon training convergence, $D(\{X,Y\})$ and $D(\{X,\hat{Y}\})$ approach 1 and 0, respectively.

% $\hat{Z}=\{X,\hat{Y}\}$ is spatio-temporally coherent with its input RTLMP (market data) video $X=\{X^1,\cdots,X^n\}$ and the ground truth video $Z=\{X,Y\}$.\\
%\indent Given ground truth $Z=\{X,Y\}$ or forecast $\hat{Z}=\{X,\hat{Y}\}$, the discriminator $D$ is trained to classify $Z=\{X,Y\}$ as real by giving scalar close to 1 and $\hat{Z}=\{X,\hat{Y}\}$ as fake by giving scalar close to 0. \\
%\indent Details of generator model, discriminator model, and adversarial training process are described below.
%\vspace{-0.3cm}
\subsection{The Discriminative Model $D$}
The discriminator $D$ is a convolutional neural network.
%followed by fully connected layers and the ReLU output\cite{mathieu2015deep,denton2015deep}. 
%$D$ takes a market data video $Z=\{X,Y\}$ or $\hat{Z}=\{X,\hat{Y}\}=\{X,G(X)\}$ as input to output a scalar $D(Z/\hat{Z})\in [0,1]$, which measures the probability that the last RTLMP image $Y$ in input video $Z=\{X,Y\}$ or $\hat{Y}$ in the input video $\hat{Z}=\{X,\hat{Y}\}$ is ground truth. 
It is trained by minimizing the following loss function (distance function) to classify the input $\{X,Y\}$ into class $1$ (i.e., $Y$ is classified as the ground-truth $X(n+1)$) and the input $\{X,\hat{Y}\}=\{X,G(X)\}$ into class $0$ (i.e., $\hat{Y}=G(X)$ is classified as the generated fake image $\hat{X}(n+1)$):
%\begin{align}
%\begin{split}
%    \mathcal{L}_{adv}^D(X,Y)=&\mathcal{L}_{bce}(D(Z),1)+\mathcal{L}_{bce}(D(\hat{Z}),0)\\
%    =&\mathcal{L}_{bce}(D(\{X,Y\}),1)\\
%    &+\mathcal{L}_{bce}(D(\{X,G(X)\}),0)
%\end{split}
%\end{align}
\begin{align}
\begin{split}
\mathcal{L}_{adv}^D(X,Y)=\mathcal{L}_{bce}(D(\{X,Y\}),1)+\mathcal{L}_{bce}(D(\{X,\hat{Y}\}),0)
\end{split}
\end{align}
%\vspace{-0.3cm}
where $\mathcal{L}_{bce}$ is the binary cross-entropy loss:
\begin{equation}
\mathcal{L}_{bce}(K,S)=-[K log(S)+(1-K)log(1-S)]
\end{equation}
%\vspace{-0.1cm}
where $K\in [0,1]$ and $S\in \{0,1\}$. $\mathcal{L}_{bce}(K,S)$ measures the distance between the discriminator output $K=D(\{X,\cdot\})$ and the label $S$ ($S_i=1$ and $S_i=0$ for real and generated images, respectively).
By minimizing the above loss function, the discriminator $D$ is forced to classify $D(\{X,Y\})$ as real (with label 1) and $D(\{X,\hat{Y}\})$ as generated (with label 0).

%when the generator $G$ is well trained to produce realistic prediction $\hat{Y}$, the discriminator $D$ cannot classify $Y$ as true and $\hat{Y}$ as fake by comparing the spatio-temporal correlations in the market data videos $\{X,Y\}$ and $\{X,\hat{Y}\}$.
%\vspace{-0.3cm}
\subsection{The Generative Model $G$}
The generator $G$ is also a convolutional nerual network, which could account for dependencies among pixels (i.e., the spatio-temporal correlations among historical market data).
%Fig.~\ref{convet} shows the basic architecture of the convolutional nerual network for next-hour RTLMP prediction.
The neural network architecture is adopted from the deep video prediction model in \cite{mathieu2015deep}.
%\begin{figure}[h]
%        \centering
%        \includegraphics[height=3.0cm,width=10.0cm]{conv.jpg} 
%        \caption{\hl{A basic convolutional nerual network for next-hour RTLMP prediction.}}
%        \label{convet}
%\end{figure}
The objective of $G$ is to minimize a certain loss function (distance function) between the generated market data image $\hat{Y}=G(X)$ and the ground-truth image $Y=X(n+1)$. This paper adopts four loss functions from \cite{mathieu2015deep}, \cite{directionref} to quantify the distances between $Y$ and $\hat{Y}$ from various perspectives. These loss functions are combined to form the following weighted loss function for training $G$:
\begin{align}
\begin{split}
\mathcal{L}^G(X,Y)=&\lambda_{adv}\mathcal{L}_{adv}^G(X,Y)+\lambda_{\ell_p}\mathcal{L}_p(X,Y)\\
&+\lambda_{gdl}\mathcal{L}_{gdl}(X,Y)+\lambda_{dcl}\mathcal{L}_{dcl}(X,Y)
\end{split}
\label{eqn_G_distance_full}
\end{align}
where $\mathcal{L}^G(X,Y)$ denotes the overall weighted loss function for training $G$; $\mathcal{L}_{adv}^G(X,Y)$, $\mathcal{L}_p(X,Y)$,  $\mathcal{L}_{gdl}(X,Y)$, and $\mathcal{L}_{dcl}(X,Y)$ denote the four individual loss function terms (introduced separately in the following sections); $\lambda_{adv}$, $\lambda_{\ell_p}$, $\lambda_{gdl}$, and $\lambda_{dcl}$ denote the hyperparameters (weights) for adjusting the tradeoffs among these loss terms. 

\subsubsection{The $p$-norm Loss Function $\mathcal{L}_p(X,Y)$}
This loss function measures the $p$-norm distance between the generated and ground-truth market data images $\hat{Y}=G(X)$ and $Y$:
\begin{equation}
\mathcal{L}_p(X,Y)=\ell_p(G(X),Y)=\left\|G(X)-Y\right\|_p^p
\label{eqn_p_norm}
\end{equation}
where $\left\|\cdot\right\|_p$ denotes the entry-wise $p$-norm of a particular matrix, with $p=1$ or $p=2$.  
Euclidean distance or Manhattan distance between $\hat{Y}$ and $Y$ is calculated when $p=2$ or $p=1$, respectively. During the training process, this loss function forces the generator $G$ to generate the next-hour market data image $\hat{Y}$ that is close to the corresponding ground-truth image $Y$ by minimizing the difference between $Y$ and $\hat{Y}$. 
%This is achieved by minimizing the difference between the ground-truth and generated market data images $Y$ and $\hat{Y}$.
\subsubsection{The Adversarial Loss Function $\mathcal{L}_{adv}^G(X,Y)$} The following loss function is adopted to capture the temporal correlations among historical market data \cite{radford2015unsupervised,mathieu2015deep,denton2015deep}: 
\begin{align}
\begin{split}
\mathcal{L}_{adv}^G(X,Y)&=\mathcal{L}_{bce}(D(\{X,G(X)\}),1)
\end{split}
\end{align}

During the training process, the adversarial loss function forces the generator $G$ to generate the next-hour market data image $\hat{Y}$ that is temporally coherent with the input historical market data video ${X}$, such that the generated video $\{X,\hat{Y}\}$ is realistic enough to confuse the discriminator $D$. This is achieved by minimizing $\mathcal{L}_{bce}(D(\{X,\hat{Y}\}),1)$, which measures the distance between the discriminator output for the generated market data video and the label for real market data video.

%\indent With the above adversarial loss function, the generator $G$ is forced to generate the next-hour market data image $\hat{Y}$ that is temporally coherent with the input historical market data video ${X}$, such that the newly-formed video $\{X,\hat{Y}\}$ is realistic enough to confuse the discriminator $D$. 

\subsubsection{The Gradient Difference Loss Function $\mathcal{L}_{gdl}(X,Y)$}
The following loss function is adopted to further capture the spatial correlations among historical market data \cite{mathieu2015deep}:
\begin{align}
\begin{split}
\mathcal{L}_{gdl}(X,Y)&=\mathcal{L}_{gdl}(\hat{Y},Y)\\
&=\sum_{i,j}\vert\vert Y_{i,j}-Y_{i-1,j}\vert-\vert\hat{Y}_{i,j}-\hat{Y}_{i-1,j}\vert \vert ^{\alpha}\\
&+\vert \vert Y_{i,j-1}-Y_{i,j}\vert-\vert\hat{Y}_{i,j-1}-\hat{Y}_{i,j}\vert\vert ^{\alpha}
\end{split}
\label{eqn_gradient_diff}
\end{align}
where $\alpha\geq1, \alpha\in\mathbb{Z}$; $Y_{i,j}$ and $\hat{Y}_{i,j}$ denote $(i,j)^{th}$ pixels in the ground-truth and generated market data images $Y$ and $\hat{Y}=G(X)$, respectively. During the training process, this loss function forces the generator $G$ to generate the next-hour market data image $\hat{Y}$ that has similar pixel gradient difference information compared to the ground-truth image $Y$. Since the pixel gradient difference information captures the spatial variations of historical market data, this minimization ensures the generated market data fully captures the spatial correlations among system-wide historical market data.

\subsubsection{The Pixel Direction Changing Loss Function $\mathcal{L}_{dcl}(X,Y)$}
Different from classical video prediction problems, we would like to correctly predict whether RTLMPs will increase or decrease in the next market clearing interval. The following loss function is introduced to capture the changing directions of market data pixels over time\cite{directionref} :
\begin{equation}
\mathcal{L}_{dcl}(X,Y)=\sum_{i,j}\vert sgn(\hat{Y}_{i,j}-X_{i,j}(t))-sgn(Y_{i,j}-X_{i,j}(t)) \vert
\label{eqn_direction}
\end{equation}
where $sgn(\cdot)$ is the sign function:

%\vspace{-0.3cm}
\begin{align}
sgn(x) & = \begin{cases}
-1 & if\ x \leq 0\\
0 & if\ x = 0\\
1 & if\ x \geq 0
\end{cases}
\end{align}
During the training process, this loss function forces the generated market data video $\{X,\hat{Y}\}$ to correctly follow the pixel changing directions in the ground-truth video $\{X,Y\}$ by penalizing incorrect market data trend predictions over time. %This is achieved by penalizing incorrect predictions of the market data trends over time.
%\vspace{-0.6cm}
\subsection{Adversarial Training}
The GAN-based RTLMP prediction model is trained through the adversarial training procedure in Fig.~\ref{training procedure}. Algorithm 1 summarizes the training algorithm. The generator $G$ and discriminator $D$ are trained simultaneously, with their model weights, $W_G$ and $W_D$, updated iteratively. The stochastic gradient descent (SGD) minimization is adopted to obtain optimal model weights. In each training iteration, a new batch of $M$ training data samples (i.e., $M$ historical market data videos) are obtained for updating $W_G$ and $W_D$. Upon convergence, the generator $G$ is trained to generate $\hat{Y}$ as realistic as possible, such that the discriminator $D$ cannot confidently classify $\{X,\hat{Y}\}$ into 0 as a generated video.
\begin{algorithm}
	\caption{Training generative adversarial networks for RTLMP prediction}
	\label{alg1}
	\begin{algorithmic}
		\REQUIRE set the learning rates $\rho_D$ and $\rho_G$, loss hyperparameters $\lambda_{adv}$, $\lambda_{\ell_p}$, $\lambda_{gdl}$, $\lambda_{dcl}$, and minibatch size M
		\REQUIRE initial discriminative model weights $W_D$ and generative model weights $W_G$
		\WHILE{not converged}
		\STATE \textbf{Update the discriminator D:}
		\STATE Get a batch of M data samples from the training dataset,
		$(X,Y)=(X^{(1)},Y^{(1)}),\cdots,(X^{(M)},Y^{(M)})$
		\STATE Do one SGD update step
		\STATE $W_D=W_D-\rho_D \sum_{i=1}^M \frac{\partial \mathcal{L}_{adv}^D(X^{(i)},Y^{(i)})}{\partial W_D}$
		\STATE \textbf{Update the generator G:}
		\STATE Get a \textit{new} batch of M data samples from the training dataset, $(X,Y)=(X^{(1)},Y^{(1)}),\cdots,(X^{(M)},Y^{(M)})$
		\STATE Do one SGD update step
		\STATE $W_G=W_G-\rho_G \sum_{i=1}^M( \lambda_{adv}\frac{\partial \mathcal{L}_{adv}^G(X^{(i)},Y^{(i)})}{\partial W_G}+\lambda_{\ell_p}\frac{\partial \mathcal{L}_{\ell_p}(X^{(i)},Y^{(i)})}{\partial W_G}+\lambda_{gdl}\frac{\partial \mathcal{L}_{gdl}(X^{(i)},Y^{(i)})}{\partial W_G}+\lambda_{dcl}\frac{\partial \mathcal{L}_{dcl}(X^{(i)},Y^{(i)})}{\partial W_G})$
		\ENDWHILE
	\end{algorithmic}
\end{algorithm} 
%\vspace{-0.3cm}
\section{Autoregressive Moving Average Calibration}
The above GAN model is trained using year-long historical market data, and applied to perform RTLMP predictions hour by hour for the following year. Due to fuel price fluctuation, load growth and generation/transmission systems upgrade, the market data statistics may vary year by year. This may cause deviations between the ground-truth and predicted RTLMPs, as the generator $G$ is trained using market data from previous years. For better prediction accuracy, the RTLMPs generated by GAN are calibrated through estimating their deviations from the ground truth:
\begin{equation}
\widetilde{y}(i+1) = \hat{y}(i+1)+{\Delta}\hat{y}(i+1)
\end{equation}
%\vspace{-0.3cm}
where
\begin{equation}
{\Delta}\hat{y}(i+1) = y(i+1) - \hat{y}(i+1) + e(i+1)
\end{equation}
where $y(i+1)$, $\hat{y}(i+1)$, and $\widetilde{y}(i+1)$ denote the ground-truth RTLMP, the RTLMP generated by $G$, and the RTLMP after calibration at time $i+1$ for a particular price node, respectively; ${\Delta}\hat{y}(i+1)$ denotes the estimated difference between $y(i+1)$ and $\hat{y}(i+1)$; $e(i+1)$ denotes the estimation error.

The ARMA model below is applied to estimate ${\Delta}\hat{y}(i+1)$:
%\vspace{-0.3cm}
\begin{align}
\begin{split}
{\Delta}\hat{y}(i+1)=&\mu+\sum_{k=1}^{p}{\phi_k}{\Delta}\hat{y}(i-k+1)\\
&+\sum_{k=1}^{q}{\theta_k}\varepsilon(i-k+1)+\varepsilon(i+1)
\end{split}
\end{align}
where $\mu$ denotes the expectation of ${\Delta}\hat{y}(i+1)$, $\phi_k$ and $\theta_k$ are the autoregressive (AR) and moving average (MA) parameters of the ARMA model, respectively; $\varepsilon(i)$ represents the white noise error terms at time $i$; $p$ and $q$ denote the orders of the AR and MA terms of the ARMA model, respectively. Appropriate values of $p$, $q$, $\mu$, $\phi_k$, $\theta_k$, and the variance of the white noise series $\varepsilon(i)$ are identified using historical data \cite{GeurtsMichael1977TSAF,PolasekWolfgang2013TSAa,MATLAB}.

\section{Feature Selection} \label{feature_selection}

Based on the data structure defined previously, each market data pixel in the GAN prediction model may contain $l$ different types of market data points (after normalization), i.e., $v_{i,j}^{norm}(t) = [v_{i,j}^{norm-1}(t), v_{i,j}^{norm-2}(t), \cdots, v_{i,j}^{norm-l}(t)]$. This section identifies the following four types of market data that are highly related to the RTLMP prediction problem and are publicly available in many US electricity markets. 
%\vspace{-0.3cm}
\subsubsection*{Real-Time LMP Data}
\begin{figure}[!h]
	\centering
	\includegraphics[width=9cm, height=3.9cm]{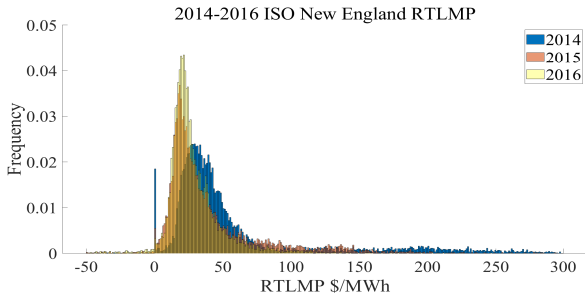} 
	\caption{Histograms generated by hourly RTLMPs from ISO-NE in 2014-2016.}
	\label{LMPdistribution}
\end{figure}

To learn the spatio-temporal correlations among historical RTLMPs, the proposed prediction model is trained using historical RTLMPs. Fig.~\ref{LMPdistribution} shows the histograms generated using RTLMPs from ISO-NE in 2014-2016. It is clear that although RTLMPs obtained during different years for the same market follow similar probability distributions, there exist discrepancies between RTLMP distributions during different years. These statistical discrepancies may decrease the RTLMP prediction accuracy if the prediction model is trained using historical RTLMPs only. Therefore, additional information is needed for training the GAN model.
%\vspace{-0.3cm}
%\vspace{-0.8cm}
\subsubsection*{Day-Ahead LMP Data}
RTLMPs tend to be strongly correlated with DALMPs, since generators committed in the day-ahead market may affect the total generation capacity and overall energy price in the real-time market. As an example, the correlation coefficient between DALMPs and RTLMPs in ISO-NE is $66.38\%$ in 2018\cite{ISONEdata}. Therefore, historical DALMPs are adopted as training inputs for the GAN model.

%\vspace{-0.5cm}
\subsubsection*{Demand Data}
%\vspace{-0.1cm}
System demand patterns and uncertainties could significantly affect RTLMPs. Since locational demand data is public in many electricity markets, these historical demands are included as training inputs for our GAN model.
%\vspace{-0.3cm}
\subsubsection*{Generation Mix Data}
%\vspace{-0.1cm}
Generation price and quantity offers play critical roles in the LMP formation process. Therefore, historical generation offer data contains important information for predicting RTLMPs. However, in most electricity markets, system-wide generation offer details are published with significant delays, and therefore cannot be fully utilized by market participants for their price prediction problem. To resolve this issue, we adopt historical hourly generation mix data that is publicly available in some markets (such as SPP \cite{SPPdata}) in training our GAN model. Table~\ref{genshare} shows the correlation coefficients calculated between SPP's historical generation mix (i.e., percentage generation by fuel type) data and historical RTLMP data \cite{SPPdata}. It is clear that the percentage generations of certain generation types are highly correlated with RTLMPs in SPP. Therefore, historical hourly generation mix data is adopted in training the RTLMP prediction model for SPP.
%These highly-correlated historical data points are therefore adopted in our GAN model for RTLMP prediction in SPP.
%\vspace{-0.3cm}
%\begin{table}[!h]
%	\centering
%	%\caption{Correlations Between Generation Mix and RTLMP in SPP} 
%	\caption{Correlations Between Generation Mix and RTLMPs in SPP}  
%	\begin{tabular}{lc}  
%		\hline 
%		\hline
%		Generation type & Correlation coefficient \\  
%		\hline  
%		Coal market  & 17.83\%\\  
%		Coal self  & 22.72\% \\  
%		Diesel fuel oil & -5.26\%\\
%		Hydro   &  19.73\%\\
%		Natural gas & 36.25\%\\
%		Nuclear & -29.34\%\\
%		Solar & 17.20\%\\
%		Waste disposal & -14.83\%\\
%		Wind  & -37.93\%\\
%		\hline
%		\hline
%	\end{tabular}
%	\label{genshare}
%\end{table}  

\begin{table}[!h]
	\centering
	%\caption{Correlations Between Generation Mix and RTLMP in SPP} 
	\caption{Correlations Between Generation Mix and RTLMPs in SPP}  
	\begin{tabular}{>{\centering\arraybackslash}p{1.7cm} >{\centering\arraybackslash}p{1.5cm} | >{\centering\arraybackslash}p{1.7cm} >{\centering\arraybackslash}p{1.5cm}}  
		\hline 
		\hline
		Generation Type & Correlation Coefficient & Generation Type & Correlation Coefficient \\  
		\hline  
		Coal market  & 17.83\% & Natural gas & 36.25\%\\  
		Coal self  & 22.72\% & Nuclear & -29.34\%\\  
		Diesel fuel oil & -5.26\% & Solar & 17.20\%\\
		Hydro   &  19.73\% & Waste disposal & -14.83\%\\
		Wind  & -37.93\%\\
		\hline
		\hline
	\end{tabular}
	\label{genshare}
\end{table}

%\vspace{-0.5cm}
\section{Case Studies}
The proposed RTLMP prediction method is tested using historical market data from ISO-NE and SPP. 
%Implementation details of the GAN model are presented below.
%\vspace{-0.3cm}
%\subsection{Implementation}
The prediction models are implemented by TensorFlow 2.0 ~\cite{tensorflow} and trained on Google Colaboratory using online GPU for acceleration. Implementation details of the GAN model are presented below.

\subsubsection*{Neural Network Architecture}
The nerual network architecture of our model is inspired by the video prediction model in \cite{denton2015deep}. Both the generative and discriminative models are deep convolutional neural networks without any pooling/subsampling layers. In the generative model, all the transpose convolutional (Conv2DTranspose) layers are followed by the batch normalization layers and ReLU units. The outputs of the generative model are normalized by a hyperbolic tangent (Tanh) function. In the discriminative model, except for the output layer, all the convolutional (Conv2D) layers and fully-connected (Dense) layers are followed by the batch normalization layers, Leaky-ReLU units, and dropout layers. Details of the nerual network architecture are listed in Table~\ref{network architecture}.
\subsubsection*{Configuration}
All the convolutional (Conv2D) and transpose convolutional (Conv2DTranspose) layers in our model are with kernel size of $3\times 3$ and stride size of $1\times 1$. The transpose convolutional (Conv2DTranspose) layers in the generative model are padded, the convolutional (Conv2D) layers in discriminative model are not padded. In the discriminative model, the dropout rates are set to 0.3, the small gradients are set to 0.2 when Leaky-ReLU is not active. All the neural networks are trained using standard SGD optimizer with a minibatch size of 4, i.e., $M=4$ in Algorithm 1. The learning rates $\rho_G$ and $\rho_D$ are set to 0.0005, without decay and momentum. The loss functions in (\ref{eqn_G_distance_full})-(\ref{eqn_direction}) are implemented with the following parameteres: $\lambda_{adv}=\lambda_{dcl}=0.2$ (in (\ref{eqn_G_distance_full})), $\lambda_{\ell_p}=\lambda_{gdl}=1$ (in (\ref{eqn_G_distance_full})), $p=2$ (in (\ref{eqn_p_norm})), and $\alpha=1$ (in (\ref{eqn_gradient_diff})). %More details on the nerual network architecture and configuration are available in [X].
%\vspace{-0.3cm}
%\begin{table}[!h]
%	\centering  
%	\caption{Nerual Network Architecture Details}  
%	\begin{tabular}{llc}  
%		\hline 
%		\hline
%		Layers & Generative Model G & Feature Maps for G\\  
%		\hline  
%		Input  & $3\times 3\times 14$ & \\  
%		Layer 1 & Conv2DTranspose & 64 \\  
%		Layer 2 & Concatenate & 896\\
%		Layer 3 & Conv2DTranspose & 1024 \\
%		Layer 4 & Conv2DTranspose & 512\\
%		Layer 5 & Conv2DTranspose & 64\\
%		Output & $3\times 3\times 1$ &\\
%		
%		\hline
%		\hline
%		Layers & Discriminative Model D & Feature Maps for D\\  
%		\hline
%		Input  & $3\times 3\times 5$ & \\  
%		Layer 1 & Conv2D & 64 \\  
%		Layer 2 & Concatenate & 320\\
%		Layer 3 & Dense & 1024\\
%		Layer 4 & Dense & 512\\
%		Layer 5 & Dense & 256\\
%		Output & scalar$\in[0,1]$ & \\
%		\hline
%		\hline
%	\end{tabular}
%	\label{network architecture}
%\end{table}   

\begin{table}[!h]
	\centering  
	\caption{Neural Network Architecture Details}  
	\begin{tabular}{>{\centering\arraybackslash}p{0.9cm} | >{\centering\arraybackslash}p{3.3cm} >{\centering\arraybackslash}p{3.3cm}}  
		\hline 
		\hline
		  & Generator G & Discriminator D \\
		  & (Layer Type, Feature Map) & (Layer Type, Feature Map)\\  
		\hline  
		Input  & $3\times 3\times 14$ & $3\times 3\times 5$ \\  
		Layer 1 & Conv2DTranspose, 64 & Conv2D, 64 \\  
		Layer 2 & Concatenate, 896 & Concatenate, 320 \\
		Layer 3 & Conv2DTranspose, 1024 &  Dense, 1024 \\
		Layer 4 & Conv2DTranspose, 512 & Dense, 512 \\
		Layer 5 & Conv2DTranspose, 64 & Dense, 256 \\
		Output & $3\times 3\times 1$ & scalar$\in[0,1]$\\
		\hline
		\hline
	\end{tabular}
	\label{network architecture}
\end{table}   

%\vspace{-0.8cm}
\subsection{Test Case Description}
The proposed approach is applied to predict zonal-level RTLMPs in ISO-NE \cite{ISONEdata} and SPP \cite{SPPdata}. For both markets, historical market data from nine price zones are organized into $3\times3$ market data images and videos. The RTLMP prediction accuracy is quantified by the mean absolute percentage error (MAPE). The test case data is described as follows.
\subsubsection{Case 1}
The training data set contains three types of hourly ISO-NE market data (zonal RTLMPs, DALMPs, demands) from 1/1/2016 to 12/31/2017. The trained model is tested by predicting ISO-NE RTLMPs hour by hour in 2018. 
%The 'System' price zone contains system-level market data averaged across ISO-NE.

\subsubsection{Case 2}
The training data set contains four types of hourly SPP market data (zonal RTLMPs, DALMPs, demands, and generation mix data) from 6/1/2016 to 7/30/2017. The model is tested by predicting SPP RTLMPs hour by hour in the following four periods: 7/31/2017-8/13/2017, 8/21/2017-9/3/2017, 9/18/2017-10/1/2017, and 10/2/2017-10/15/2017.
%The 'NHub' and 'SHub' price zones contain hub-level market data averaged across the north and south hubs of SPP, respectively.

It should be noted that although the case studies in this section are preformed using zonal LMPs, the proposed method can be easily applied to predicting nodal LMPs if the model is trained using nodal-level market data.
%The mean absolute percentage error (MAPE) is adopted to quantify the RTLMP prediction errors.
%%\vspace{-0.3cm}
%\begin{equation}
%MAPE=\frac{1}{n}\sum_{i=1}^n \vert \frac{\hat{y}_i-y_i}{y_i}\vert \times 100\%
%\end{equation}
%%\vspace{-0.1cm}
%where $\hat{y}_i$ is the predicted RTLMP, $y_i$ is the ground truth of RTLMP, $n$ is the number of tested data points. 
%\vspace{-0.5cm}
\subsection{Performance Analysis}
%\vspace{-0.1cm}
Table~\ref{accuracy in case 1} shows the RTLMP prediction accuracy of our proposed approach in Case 1. For the ISO-NE test case, the annual MAPEs in 2018 are around $11\%$ for all the nine price zones. There are other works predicting RTLMPs in ISO-NE \cite{onpeakavene,simulatednedata}. In \cite{onpeakavene}, a weekly MAPE of $10.87\%$ is achieved by predicting daily average on-peak-hour prices in ISO-NE. Since the daily average on-peak-hour price is the daily averaged price from hour 8 to hour 23, this averaged price is expected to have much smoother behavior and less spikes compared to the hourly RTLMPs in our test case. When tested on raw hourly RTLMPs for an entire year without averaging/smoothing over the testing data, our approach manages to achieve reasonable accuracy compared to \cite{onpeakavene}. When tested on raw hourly RTLMPs obtained for the same week as \cite{onpeakavene} at a different year, our approach achieves a weekly MAPE of $9.08\%$, which outperforms the work in \cite{onpeakavene} using averaged prices. In \cite{simulatednedata}, an average MAPE of $10.81\%$ is achieved by predicting ISO-NE prices at four test weeks in March, June, September, and December. However, the testing data in \cite{simulatednedata} is generated through simulations. The simulation rules assume all possible explicit price-responsive behavior is known, which is not practical in real-world price prediction. Our proposed approach, when tested using year-long actual market data, has very similar MAPEs compared to \cite{simulatednedata}.
%\vspace{-0.3cm}
\begin{table}[!h] 
	\centering  
	\caption{Zonal RTLMP Prediction Accuracy in Case 1}  
	\begin{tabular}{p{1.3cm} |p{0.35cm} p{0.35cm} p{0.35cm} p{0.35cm} p{0.35cm} p{0.35cm} p{0.35cm} p{0.35cm} p{0.45cm}  }
		\hline 
		\hline
		Price Zone & VT & HN & ME & WC MA & Sys-tem & NE MA & CT & RI & SE MA \\
		\hline
		MAPE (\%) & 11.03 & 11.25 & 11.82 & 10.99 & 11.06 & 11.05 & 11.04 & 11.01 & 11.05\\
		
		\hline
		\hline
	\end{tabular}
	\label{accuracy in case 1}
\end{table}   
%\vspace{-0.3cm}

In \cite{onehouraheadjapan,onehouraheadspanish} similar hour-by-hour price prediction approaches are tested using public data from Japan market (with an price prediction MAPE of $14.28\%$) and Spanish market (with an price prediction MAPE of $15.83\%$), respectively. Our approach has much lower MAPEs compared to these works.

%\vspace{-0.3cm}
%\begin{table}[!h] 
%	\centering  
%	\begin{threeparttable}
%		%\centering  
%		\caption{RTLMP Prediction Accuracy in Case 2 and \cite{8733097}}  
%		\begin{tabular}{>{\centering\arraybackslash}p{1.5cm}|>{\centering\arraybackslash}p{2.5cm} >{\centering\arraybackslash}p{2.5cm}}  
%			\hline 
%			\hline
%			Price Zone & Approach & MAPE (\%) \\  
%			\hline
%			SHub & ALG+$\hat{M}$\tnote{1} & 25.4 \\
%			& Genscape\tnote{2} & 21.7 \\
%			& Case 2  &  17.7 \\
%			\hline
%			NHub & ALG+$\hat{M}$ & 36.9 \\
%			& Genscape & 28.2 \\
%			& Case 2  &  19.1 \\
%			
%			\hline
%			\hline
%		\end{tabular}
%		\begin{tablenotes}
%			\footnotesize
%			\item[1] The proposed method with best performance in \cite{8733097}
%			\item[2] State of art baseline prediction purchased from Genscape\cite{8733097}
%		\end{tablenotes}
%		\label{accuracy case 2}
%	\end{threeparttable}
%\end{table}   
%\vspace{-0.3cm}

%Because SPP has more market participants and a larger market territory compared to ISO-NE, the RTLMP prediction problem becomes more complex in SPP, and the prediction accuracy in Case 2 is lower than that in Case 1. \cite{8733097} did day ahead RTLMP prediction of 'SHub' and 'NHub' in SPP for same testing windows. Table \ref{accuracy case 2} illustrates the comparison of prediction accuracy between Case 1 and \cite{8733097}. The proposed method achieves much better accuracy than the baseline purchased from commercial tool which takes more confidential information as inputs.

\begin{figure}[!h]
	\centering
	\includegraphics[width=9.0cm, height=2.5cm]{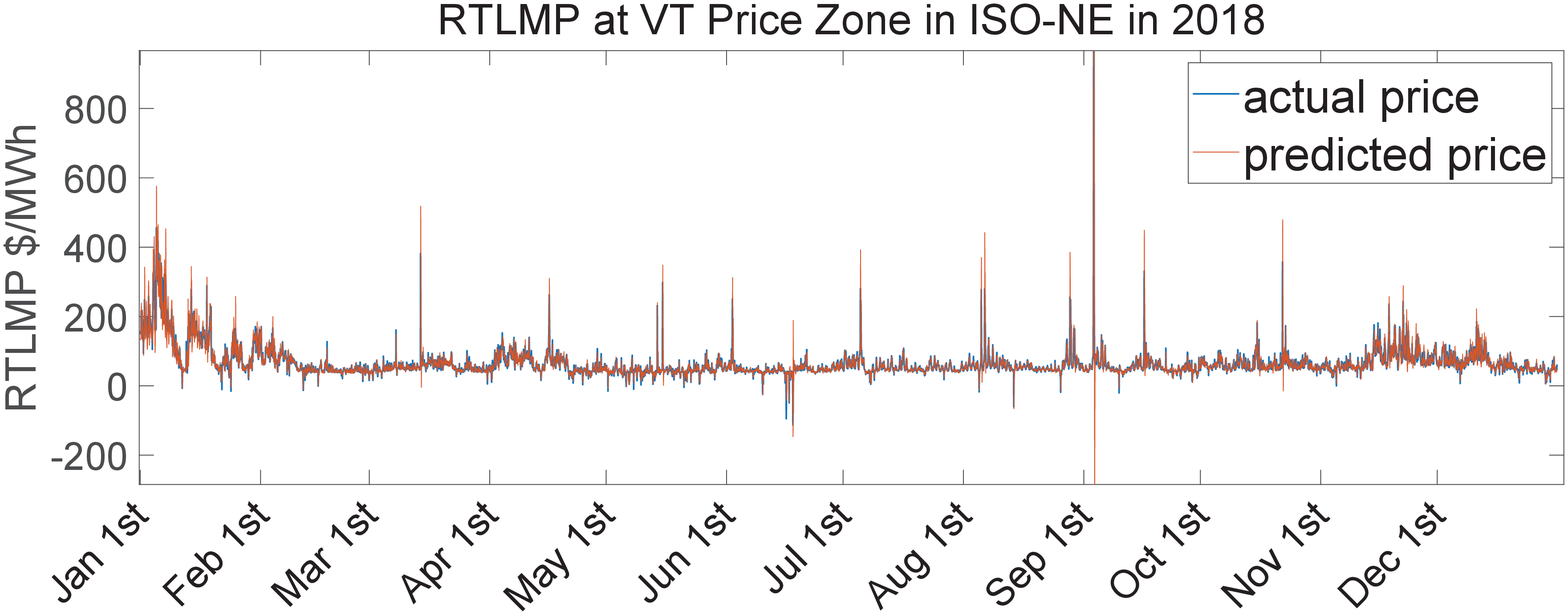} 
	\caption{Ground-truth and predicted RTLMPs for VT price zone in Case 1.}
	\label{case2}
\end{figure}

Fig.~\ref{case2} shows the ground-truth and predicted RTLMPs for VT price zone in ISO-NE during the entire year of 2018. It is clear our predicted RTLMPs closely follow the overall trends of the ground-truth RTLMPs, and successfully capture most price spikes in the testing window.

Table~\ref{accuracy case 2} shows the RTLMP prediction accuracy of our proposed approach in Case 2 and the MAPEs obtained using two other approaches in \cite{8733097}. The three approaches are tested using the same testing data sets from the SHub and NHub price zones in SPP. Compared to existing works, our proposed approach has better prediction accuracies for both price zones. It is worth mentioning that the prediction accuracies in Case 2 (SPP) are lower than those in Case 1 (ISO-NE). This is because SPP has larger market territory and more market participants compared to ISO-NE. These facts lead to more complex and harder-to-predict market dynamics in SPP.

\begin{table}[!h] 
	\centering  
	\begin{threeparttable}
		%\centering  
		\caption{RTLMP Prediction Accuracy in Case 2 and \cite{8733097}}  
		\begin{tabular}{>{\centering\arraybackslash}p{1.5cm}|>{\centering\arraybackslash}p{2.5cm} >{\centering\arraybackslash}p{2.5cm}}  
			\hline 
			\hline
			Approach & MAPE (\%) for & MAPE (\%) for \\
				     & SHub Price Zone & NHub Price Zone \\    
			\hline
			ALG+$\hat{M}$\tnote{1} & 25.4 & 36.9\\
			Genscape\tnote{2} & 21.7 & 28.2 \\
			Case 2  &  17.7 &  19.1 \\
			\hline
			\hline
		\end{tabular}
		\begin{tablenotes}
			\footnotesize
			\item[1] The proposed method with the best performance in \cite{8733097}
			\item[2] State of art baseline prediction from Genscape\cite{8733097}
		\end{tablenotes}
		\label{accuracy case 2}
	\end{threeparttable}
\end{table}  

%Fig.~\ref{sppplot} demonstrates our prediction closely follow the trend of ground truth RTLMP and also accurately represents the daily and weekly characteristic of the RTLMPs. Comparing the prediction of price zone SHub and price zone CSWS shown in Fig.~\ref{sppplot}, the proposed method accurately predict the spike appearing in one specific zone at the ending of August 21, which can be conclude that our proposed method successfully captures the spatio-temporal correlation across wholesale electricity market. However, there are several spikes which are predicted with their magnitude much lower than the ground-truth value. The failure of spike prediction contributes significantly to the relatively large MAPE.

\begin{figure}[h]
	\centering
	\includegraphics[width=8.0cm, height=3cm]{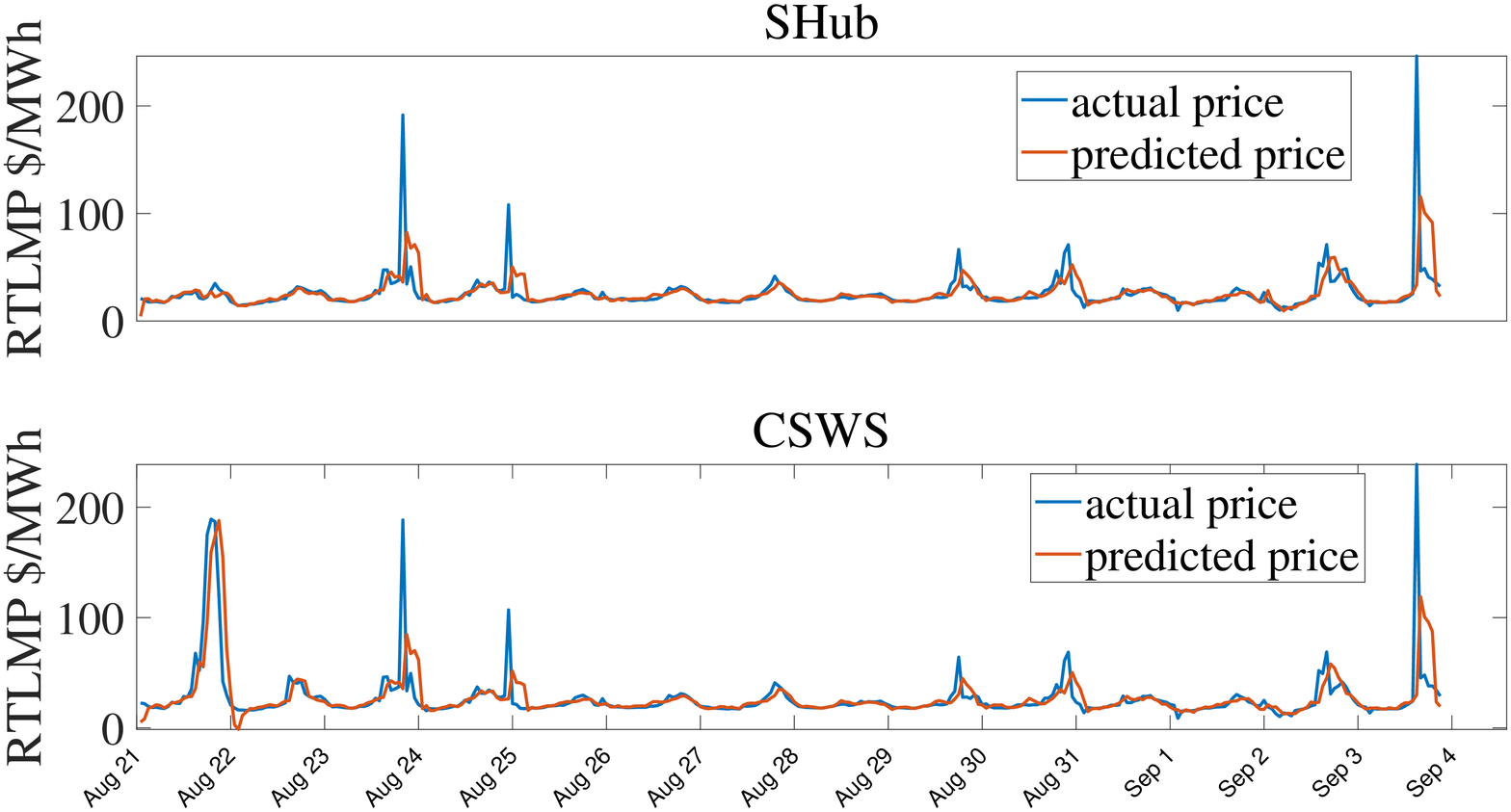} 
	\caption{Ground-truth and predicted RTLMPs for SHub and CSWS price zones in Case 2.}
	\label{sppplot}
\end{figure}

Fig.~\ref{sppplot} shows the ground-truth and predicted RTLMPs during the testing window of 8/21/2017-9/3/2017, for the SHub and CSWS price zones in SPP. It is clear that our predictions closely follow the ground-truth RTLMPs and accurately represent the daily and weekly RTLMP variations.
%\vspace{-0.3cm}
The proposed approach also successfully captures several price spikes during Aug 21, Aug 29, Aug 30, and Sep 2. It should be noted that the price spike during Aug 21 appears only in the CSWS price zone (not in the SHub price zone). This spatial characteristic is accurately reflected in the predicted zonal-level RTLMPs. It indicates our proposed approach can learn the spatio-temporal correlations among system-wide RTLMPs successfully. For the price spikes during Aug 23, Aug 25, and Sep 3, although the proposed approach fails to predict the magnitudes of the spikes, it does predict the price changing directions and the shapes of the spikes accurately. The mismatches between the ground-truth and predicted price spike magnitudes contribute significantly to the MAPEs in Case 2.

\begin{figure}[htbp]
	\centering
	\includegraphics[width=6.0cm,height=2.2cm]{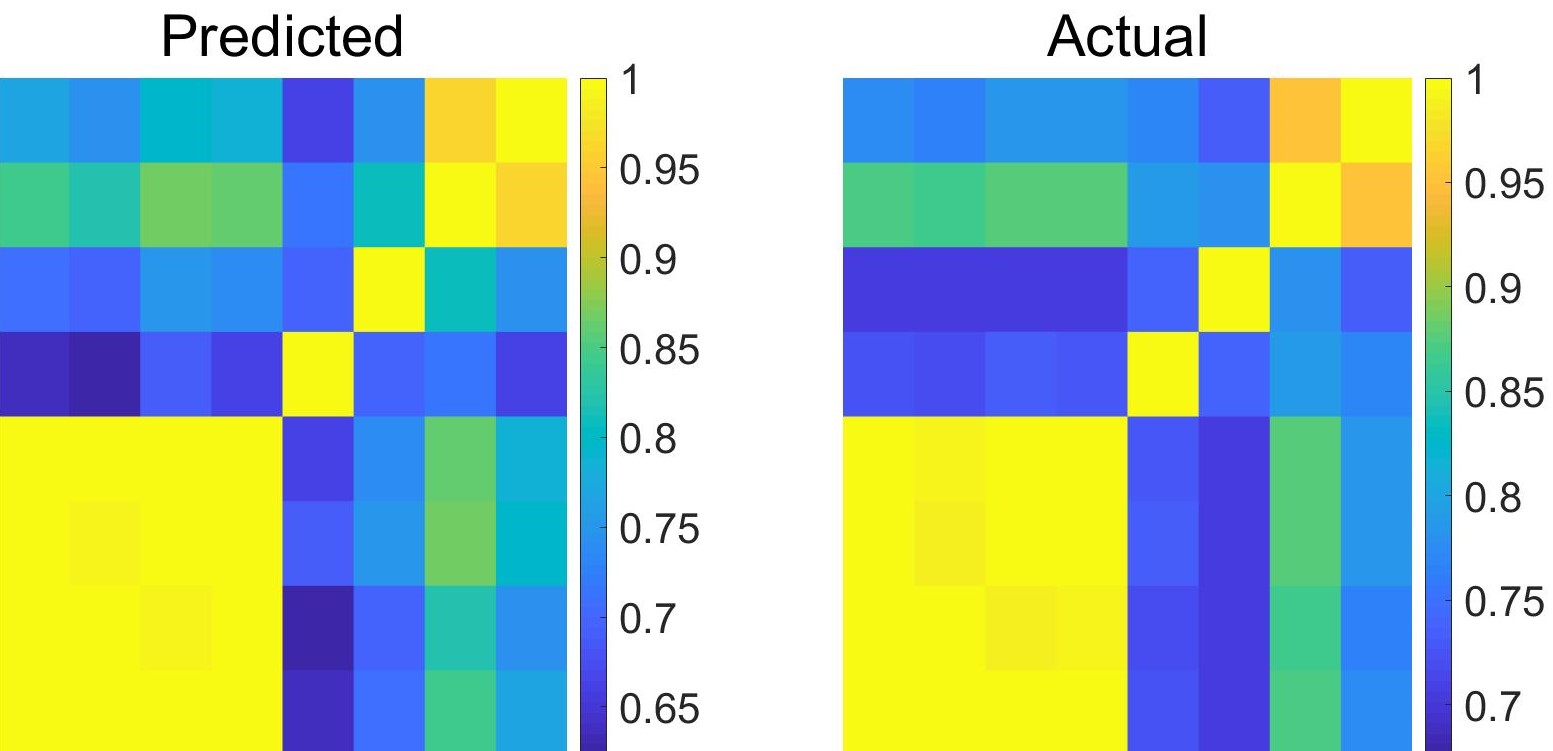} 
	\caption{The spatial correlation coefficients matrix heatmap generated using predicted RTLMPs (left) and ground-truth RTLMPs (right) in Case 2.}
	\label{spatiocase2}
\end{figure}

%To further demonstrate that our proposed method can capture the spatial correlation among price nodes over whole market territory, 
Fig.~\ref{spatiocase2} compares the spatial correlations obtained using predicted RTLMPs with those obtained using ground-truth RTLMPs in Case 2. Each $9\times 9$ heatmap matrix in Fig.~\ref{spatiocase2} visualizes the spatial correlation coefficients among 9 price zones in SPP. It is clear our RTLMP prediction model successfully captures the spatial correlations across SPP market.

%\vspace{-0.3cm}
%\vspace{-0.3cm}

\section{Conclusions and Future Work}
In this paper, a GAN-based approach is proposed to predict system-wide RTLMPs. Inspired by advanced video prediction models, the proposed method organizes historical market data into the format of images and videos. The RTLMP prediction problem is then formulated as a video prediction problem and solved using the proposed deep convolutional GAN model with multiple loss functions and an ARMA calibration process. Case studies using real-world historical market data from ISO-NE and SPP verify the performance of the proposed approach. Future work could focus on price spikes prediction by incorporating weather and public contingency data.

%Built upon this work, future work could focus on incorporating public contingency data and weather data into the proposed framework for price spikes prediction.

%\section{Conclusion}
%The conclusion goes here.
%
%
%
%
%
%% if have a single appendix:
%%\appendix[Proof of the Zonklar Equations]
%% or
%%\appendix  % for no appendix heading
%% do not use \section anymore after \appendix, only \section*
%% is possibly needed
%
%% use appendices with more than one appendix
%% then use \section to start each appendix
%% you must declare a \section before using any
%% \subsection or using \label (\appendices by itself
%% starts a section numbered zero.)
%%
%
%
%\appendices
%% you can choose not to have a title for an appendix
%% if you want by leaving the argument blank
%\section{}
%% use section* for acknowledgment
%\section*{Acknowledgment}

\ifCLASSOPTIONcaptionsoff
  \newpage
\fi



%\bibliographystyle{IEEEtran}
%\begin{thebibliography}{1}

\bibliographystyle{IEEEtran}
\bibliography{IEEEabrv,reference}


%\end{thebibliography}


%\end{IEEEbiography}
%% insert where needed to balance the two columns on the last page with
%%\newpage


%\vfill


\end{document}